\documentclass[conference]{IEEEtran}
\IEEEoverridecommandlockouts

\usepackage{cite}
\usepackage{amsmath,amssymb,amsfonts}
\usepackage{algorithm,algpseudocode}
\usepackage{graphicx}
\usepackage{textcomp}
\usepackage{xcolor}
\usepackage{makecell}
\usepackage{multirow}
\usepackage{url}
\usepackage{amsthm}
\usepackage[inline]{enumitem}
\usepackage{booktabs}
\usepackage{atbegshi}
\usepackage{eso-pic}
\def\BibTeX{{\rm B\kern-.05em{\sc i\kern-.025em b}\kern-.08em
    T\kern-.1667em\lower.7ex\hbox{E}\kern-.125emX}}

\newtheorem{proposition}{Proposition}
\theoremstyle{definition}
\newtheorem{definition}{Definition}

\begin{document}
\AddToShipoutPictureBG*{\AtPageLowerLeft{\raisebox{3.2\baselineskip}{\parbox{\paperwidth}{\centering\normalsize\textit{Accepted at WI-IAT 2025}}}}}

\algrenewcommand\algorithmicrequire{\textbf{Input:}}
\algrenewcommand\algorithmicensure{\textbf{Output:}}

\title{Pruning Overparameterized Multi-Task Networks for Degraded Web Image Restoration}

\author{\IEEEauthorblockN{Thomas Katraouras}
\IEEEauthorblockA{\textit{Dept. of Electrical and Computer Engineering} \\
\textit{University of Thessaly}\\
Volos, Greece \\
tkatraouras@uth.gr}
\and
\IEEEauthorblockN{Dimitrios Rafailidis}
\IEEEauthorblockA{\textit{Dept. of Electrical and Computer Engineering} \\
\textit{University of Thessaly}\\
Volos, Greece \\
draf@uth.gr}
}

\maketitle

\begin{abstract}
Image quality is a critical factor in delivering visually appealing content on web platforms. However, images often suffer from degradation due to lossy operations applied by online social networks (OSNs), negatively affecting user experience. Image restoration is the process of recovering a clean high-quality image from a given degraded input. Recently, multi-task (all-in-one) image restoration models have gained significant attention, due to their ability to simultaneously handle different types of image degradations. However, these models often come with an excessively high number of trainable parameters, making them computationally inefficient. In this paper, we propose a strategy for compressing multi-task image restoration models. We aim to discover highly sparse subnetworks within overparameterized deep models that can match or even surpass the performance of their dense counterparts. The proposed model, namely MIR-L, utilizes an iterative pruning strategy that removes low-magnitude weights across multiple rounds, while resetting the remaining weights to their original initialization. This iterative process is important for the multi-task image restoration model’s optimization, effectively uncovering “winning tickets” that maintain or exceed state-of-the-art performance at high sparsity levels. Experimental evaluation on benchmark datasets for the deraining, dehazing, and denoising tasks shows that MIR-L retains only 10\% of the trainable parameters while maintaining high image restoration performance. Our code, datasets and pre-trained models are made publicly available at https://github.com/Thomkat/MIR-L.
\end{abstract}


\section{Introduction}
Image quality is a critical factor in delivering visually appealing content across web platforms, where images are essential to user engagement and experience. However, images on the web frequently undergo lossy operations applied by online social networks (OSNs), such as JPEG compression and format conversion~\cite{osn_compression_1,osn_compression_2,osn_compression_3,osn_compression_4}. These operations result in noticeable degradation, with higher compression ratios correlated with greater degradation~\cite{image_compression}. Such reductions in image quality can negatively impact user experience, as lower visual quality reduces the perceived value of online content~\cite{compression_ux}. Overcoming the problem of degraded images is important for improving user experience on the web.

Image restoration is a fundamental task in computer vision that aims to recover high-quality images from degraded versions. This degradation can be caused by factors such as noise~\cite{DnCNN_Denoise}, rain~\cite{MSPFN_Derain}, haze~\cite{AODNet_Dehaze}, motion blur~\cite{MSSNet_Deblur}, low resolution~\cite{Srdiff_SISR}, or compression artifacts~\cite{artifacts_related}. Image restoration seeks to enhance the visual quality and clarity of images, making them more suitable for various applications. Recent image restoration models utilize deep learning techniques~\cite{MIMO-UNET_Deblur,DL_AIO,MSPFN_Derain,MSSNet_Deblur,AODNet_Dehaze,ESRT_SISR,EPDN_Dehaze,BRDNet_Denoise,SIRR_Derain,Restormer,FFDNet_Denoise,CVANet_SISR} to reconstruct clean images by learning complex mappings from degraded inputs to their high-quality equivalents. These models offer breakthrough performance compared to conventional restoration methods~\cite{Deblur_SR_traditional,Dehaze_traditional,denoise_traditional,SR_traditional} and are widely used in fields such as medical imaging~\cite{medicalimaging_related1}, astronomy~\cite{astronomy_related1} and aerial imaging~\cite{aerialimaging_related1}, where accurate and visually enhanced images are essential for analysis and decision-making. In addition, image restoration not only improves visual fidelity, but it also promotes high performance in tasks such as object detection~\cite{objectdetection_related2,objectdetection_related1}.

Image restoration models have been designed to handle specific tasks, such as denoising~\cite{BRDNet_Denoise,DnCNN_Denoise,FFDNet_Denoise}, deraining~\cite{MSPFN_Derain,SIRR_Derain,UMRL_Derain}, dehazing~\cite{FDGAN_Dehaze,AODNet_Dehaze,EPDN_Dehaze}, deblurring~\cite{MIMO-UNET_Deblur,RSST_Deblur,MSSNet_Deblur} and super-resolution~\cite{Srdiff_SISR,ESRT_SISR,CVANet_SISR}. However, real-world images often suffer from multiple types of degradation. To address this, the focus has shifted towards multi-task (all-in-one) image restoration models~\cite{AdaIR,DL_AIO,AirNet,PromptIR}, handling various types of degradation within a single framework, without requiring any prior knowledge of the degradation. Multi-task models provide an efficient and unified solution for real-world image restoration problems, as they reduce the overhead of deploying separate networks for individual degradations.

However, despite their effectiveness, multi-task image restoration models often require a significantly high number of trainable parameters, as demonstrated in our experiments in Section \ref{sec:Experiments}. This leads to substantial computational costs and memory demands, making running these models feasible only on high-end machines, rather than consumer-grade devices. Additionally, this limits their usability in real-time applications, such as client-side web image restoration. To address the complexity issue, researchers have explored several techniques to reduce the size and computational requirements of deep learning models, while maintaining their performances. Model compression methods such as one-shot pruning~\cite{oneshot_related1,oneshot_related2,oneshot_related3}, knowledge distillation~\cite{distillation_related3,distillation_related1,distillation_related2} and parameter sharing~\cite{parametersharing_related2,parametersharing_related1} have been applied to deep neural networks. One-shot pruning removes redundant parameters in a single step, knowledge distillation transfers knowledge from a large teacher model to a smaller student model and parameter sharing reduces redundancy by reusing parameters across different tasks or layers, effectively lowering the model size. These techniques have shown promising results in reducing the size of complex deep learning models. However, achieving a balance between preserving the model's ability to handle diverse degradations in the image restoration problem and minimizing redundant parameters still remains a challenge.

One promising approach to model compression is the Lottery Ticket Hypothesis (LTH), which suggests that within a large, overparameterized neural network, there are small subnetworks—referred to as "winning tickets"—that can be trained in isolation to achieve comparable performance to the original model~\cite{LTH}. It has been studied in image classification~\cite{LTH_Classification1,LTH_Classification2} and natural language processing~\cite{LTH_NLP2,LTH_NLP1}. The effectiveness of lottery tickets in reducing the size of multi-task image restoration models has not been thoroughly explored. Investigating LTH in this context could reveal whether certain subnetworks consistently perform well across multiple image restoration tasks, potentially enabling more efficient all-in-one solutions for handling diverse image degradations.

In this paper we propose the MIR-L model based on lottery tickets for compressing multi-task image restoration models while maintaining the performance high. Specifically, we make the following contributions:
\begin{itemize}
    \item We propose a LTH-based pruning algorithm designed for multi-task image restoration models, focusing on deraining, dehazing and denoising tasks. The algorithm iteratively removes the smallest-magnitude weights and resets the remaining weights to their original initialization, seamlessly integrating to the multi-task image restoration models' optimization process.
    \item We explore both layer-wise and global pruning strategies to assess their effectiveness in discovering sparse networks. We show that global pruning is capable of finding very sparse winning tickets, while layer-wise pruning diminishes performance.
\end{itemize}

We conduct experiments on benchmark datasets for the deraining, dehazing, and denoising tasks, comparing our proposed MIR-L with baseline pruning methods and state-of-the-art multi-task image restoration models. Our results demonstrate that the sparse networks of MIR-L reduce the number of trainable parameters by up to 90\% compared to the original dense models and outperform baseline pruning methods. In many cases, these sparse networks match or even exceed the performance of dense, state-of-the-art multi-task image restoration models, confirming that our approach effectively discovers efficient and highly sparse subnetworks—winning tickets—for multi-task image restoration.

The remainder of this paper is structured as follows: Section \ref{sec:Preliminaries} provides an overview of pruning techniques. Section \ref{sec:Proposed Method} presents our proposed method, the architecture of the multi-task image restoration model, the pruning strategy and the MIR-L optimization algorithm. Section \ref{sec:Experiments} provides the experimental evaluation, showcasing results on various datasets. Finally, Section \ref{sec:Conclusion} concludes the paper, summarizing key findings and discussing potential future directions.

\section{Preliminaries}\label{sec:Preliminaries}
Pruning is a technique in deep learning used to reduce the number of parameters in a neural network by removing certain connections. The goal is to create an efficient model with reduced memory and computational costs, while preserving the performance high. Formally, given a dense network \( f(x; \theta) \), pruning identifies and removes a subset of parameters \( \theta_p \subset \theta \), yielding a sparse network \( f(x; \theta \setminus \theta_p) \)~\cite{OptimalBrainDamage}.
Below, we outline preliminaries of existing pruning strategies.

\subsubsection{Magnitude Pruning.} 
Magnitude-based pruning is a widely used pruning strategy that removes parameters having the smallest absolute values, assuming they contribute less to the network's performance and can be removed with minimal impact. Formally, given a trained dense network \( f(x; \theta) \) and a threshold \( \tau \), a parameter \( \theta_i \) is pruned if \( |\theta_i| < \tau \), setting \( \theta_i = 0 \) for such parameters. The resulting pruned network is represented as \( f(x; \theta') \), where \( \theta' = \theta \setminus \{\theta_i : |\theta_i| < \tau\} \)~\cite{magnituedprune_related2,magnituedprune_related1}.

\subsubsection{One-shot Pruning.} 
One-shot pruning is a pruning strategy where the network parameters are pruned once after the initial training phase. A fixed percentage \( p\% \) of the parameters are removed based on a pruning criterion, e.g., magnitude-based, resulting in a sparse network with pruned parameters set to zero~\cite{oneshot_related1,oneshot_related2}.

\subsubsection{Iterative Pruning}
Iterative pruning is an approach to network sparsification, where pruning is performed in multiple rounds rather than in a single step. This method iteratively prunes a percentage \( p\% \) of the parameters and optimizes the network after each pruning step~\cite{iterativeprune_related1,iterativeprune_related2}.

\section{Proposed Method}\label{sec:Proposed Method}
\subsection{Multi-Task Image Restoration Model}\label{sec:RestorationModel}
\subsubsection{Tasks}
A multi-task (all-in-one) blind image restoration model is designed to recover clean images from degraded inputs without prior knowledge of the degradation type. Specifically, it handles the following image restoration tasks: I. \textit{Deraining}: removes rain streaks and artifacts; II. \textit{Dehazing}: removes haze and fog; III. \textit{Denoising}: reduces unwanted noise caused by low-light conditions, sensor imperfections, or compression artifacts.

\subsubsection{Architecture}
A multi-task image restoration model takes a degraded image \( \mathbf{\tilde{I}} \in \mathbb{R}^{H \times W \times C} \) as input, where \( H \times W \) is the spatial resolution, and \( C = 3 \) represents the RGB color channels~\cite{PromptIR}. This image has undergone an unknown degradation \(\mathbf{D}\). The model produces a restored image \( \mathbf{I} \in \mathbb{R}^{H \times W \times C} \). The model follows a UNet-style network architecture~\cite{Restormer} with transformer blocks~\cite{AttentionIsAllYouNeed} in both the encoding and decoding stages. Initially, low-level features \( \mathbf{F}_0 \in \mathbb{R}^{H \times W \times C} \) are extracted from \(\mathbf{\tilde{I}}\) by applying a \(3 \times 3\) convolution: \(\mathbf{F}_0 = \mathrm{Conv}_{3\times3}(\mathbf{\tilde{I}})\). These features go through a four-level hierarchical encoder-decoder, where each level increases channel capacity while reducing spatial resolution, ultimately generating low-resolution latent features \( \mathbf{F}_l \in \mathbb{R}^{\frac{H}{8} \times \frac{W}{8} \times 8C} \)~\cite{PromptIR}. The decoder gradually upsamples and refines \(\mathbf{F}_l\), leading to the final clean output image \(\mathbf{I}\). During decoding, the model incorporates sequential prompt blocks at multiple levels to inject degradation-aware information. Each prompt block consists of two components: a Prompt Generation Module (PGM) and a Prompt Interaction Module (PIM). Given $N$ learnable prompt components \(\mathbf{P}_c \in \mathbb{R}^{N \times \hat{H} \times \hat{W} \times \hat{C}}\) and input features \(\mathbf{F}_l \in \mathbb{R}^{\hat{H} \times \hat{W} \times \hat{C}}\), the prompt block produces refined features \(\mathbf{\hat{F}_l} = \text{PIM}(\text{PGM}(\mathbf{P_c}, \mathbf{F_l}), \mathbf{F_l}) \). The PGM learns an adaptive prompt $\mathbf{P}$ conditioned on both \(\mathbf{F}_l\) and \(\mathbf{P}_c\). In particular, the PGM aggregates spatial information from \( \mathbf{F}_l \) using global average pooling, followed by a \(1 \times 1\) convolution and softmax to produce prompt weights: \(\mathbf{w} = \mathrm{Softmax}(\mathrm{Conv}_{1\times1}(\mathrm{GAP}(\mathbf{F}_l)))\), where \(\mathbf{w} \in \mathbb{R}^{1 \times 1 \times N}\). These weights \(\mathbf{w}\) determine the contribution of each prompt component \(\{\mathbf{P}_{c_1}, \dots, \mathbf{P}_{c_N}\}\) in a weighted sum. The resulting combination is then refined by a \(3 \times 3\) convolution: \(\mathbf{P} = \mathrm{Conv}_{3\times3}(\sum_{i=1}^N w_i \, \mathbf{P}_{c_i})\). The PIM fuses \(\mathbf{P}\) with \(\mathbf{F}_l\) by concatenating along the channel dimension: \(\mathbf{F}_\mathrm{concat} = \mathrm{Concat}(\mathbf{F}_l, \mathbf{P})\). A transformer block $\mathbf{T}$ processes \(\mathbf{F}_\mathrm{concat}\) to incorporate degradation-specific information, followed by two consecutive \(1 \times 1\) and \(3 \times 3\) convolutions: \(\mathbf{\hat{F}_l} = \mathrm{Conv}_{3\times3}\!\bigl(\mathrm{Conv}_{1\times1}(\mathbf{T}(\mathbf{F}_\mathrm{concat}))\bigr)\). Finally, \( \mathbf{\hat{F}_l} \) propagate through the decoder, leading to the reconstructed image \( \mathbf{I} \). The \(L_1\) loss function is used to minimize the absolute differences between the restored and ground truth images, defined as \(L_1 = \frac{1}{HWC} \sum_{i=1}^{HWC} \left| \mathbf{I_{\text{GT}}}_i - \mathbf{I}_i \right|\), where \( \mathbf{I_{\text{GT}}} \in \mathbb{R}^{H \times W \times C} \) is the ground truth image~\cite{PromptIR}. The optimization is performed using the Adam optimizer.

\subsection{Lottery Ticket Hypothesis}
The LTH proposes that within a dense, randomly-initialized neural network, there is a sparse subnetwork—referred to as a winning ticket—that can be trained in isolation to achieve performance comparable to the original network~\cite{LTH}.

\begin{definition}[Winning Ticket]
A winning ticket, denoted as \( f_w(x; \theta_w) \), is a sparse subnetwork within a dense, randomly-initialized neural network \( f(x; \theta) \) with initial parameters \( \theta_0 \sim \mathcal{D}_\theta \), such that when trained in isolation from its original initialization \( \theta_0 \), it satisfies \( a_{f_w} \geq a_f \), \( j_{f_w} \leq j_f \), and \( |\theta_w| \ll |\theta| \); where \( a_{f_w} \) and \( a_f \) denote the test accuracies achieved by \( f_w \) and \( f \), respectively, \( j_{f_w} \) and \( j_f \) denote the number of training iterations required to reach minimum validation loss, and \( |\theta_w| \) and \( |\theta| \) denote the number of parameters in the winning ticket and the original network, respectively.
\end{definition}

\begin{proposition}
Consider a dense feed-forward neural network \( f(x; \theta) \) with initial parameters \( \theta_0 \sim \mathcal{D}_\theta \). Let \( m \in \{0, 1\}^{|\theta|} \) be a binary mask that identifies the active connections in the subnetwork. The Lottery Ticket Hypothesis predicts that a mask \( m \) does exist such that training \( f(x; m \odot \theta_0) \), where \( \odot \) denotes element-wise multiplication, results in a winning ticket \( f_w(x; \theta_w) \); where \(\theta_w \equiv m \odot \theta_0\).
\end{proposition}

\subsubsection{Layer-wise Pruning}
Layer-wise pruning is a strategy where pruning is applied independently to each layer of the network. A fixed percentage \( p\% \) of the smallest-magnitude weights within each layer are pruned, ensuring that sparsity is uniformly distributed across all layers. The output layer is pruned at half the rate, \( \frac{p}{2}\% \), since it typically contains far fewer parameters compared to other layers. Pruning it too aggressively can lead to diminishing returns much earlier.

\subsubsection{Global Pruning}
Global pruning is a strategy where a fixed percentage \( p\% \) of the smallest-magnitude weights are pruned across the entire network, rather than on a per-layer basis. This approach is particularly effective in deeper networks, where layers can have significantly different numbers of parameters. By pruning globally, bottlenecks caused by uniformly pruning smaller layers are avoided. As a result, global pruning can identify smaller winning tickets compared to layer-wise pruning, especially in networks with imbalanced layer sizes.

\begin{algorithm}[t]
\caption{MIR-L Optimization Algorithm}
\label{alg:promptir_lth}
\begin{algorithmic}[1]
\small
\Require
  \begin{enumerate*}[label=\arabic*.]
    \item Initial dense network \(f(\mathbf{\tilde{I}}; \theta_0)\),
    \item Pruning rate \(p\),
    \item Number of training epochs \(j\),
    \item Number of warmup epochs \(j_w\),
    \item Batch size \(B\),
    \item Training samples \(\mathcal{X}_{train}\),
    \item Initial learning rate \(\eta_{\text{start}}\),
    \item Base learning rate \(\eta_{\text{base}}\),
    \item Minimum learning rate \(\eta_{\text{min}}\),
    \item Target sparsity level \(S\)
  \end{enumerate*}
\vspace{0.2em}
\Ensure Trained sparse network \( f(\mathbf{\tilde{I}}; \theta) \)
\vspace{0.2em}
\State \( m \gets \mathbf{1}^{|\theta_0|} \) \Comment{Initialize binary mask}
\State \( \theta \gets m \odot \theta_0 \) \Comment{Set initial parameters}

\While{\(\tfrac{\| m \|_0}{|\theta_0|} < S\)}
    \For{epoch = 1 \(\to\) \(j\)}
        \State Compute learning rate \(\eta_t\) using Linear Warmup Cosine Annealing:
        \[
        \eta_t =
        \begin{cases}
            \eta_{\text{start}} + \frac{t}{j_w - 1} (\eta_{\text{base}} - \eta_{\text{start}}), & 0 \leq t < j_{w} \\
            \eta_{\text{min}} + \frac{1}{2} (\eta_{\text{base}} - \eta_{\text{min}}) \left(1 + \cos\left(\frac{(t - j_w) \pi}{j - j_w} \right) \right), & j_w \leq t \leq j
        \end{cases}
        \]

        \For{step = 1 \(\to\) \(\frac{|\mathcal{X}_{train}|}{B}\)}
        \vspace{3pt}
            \State \( \mathbf{I}_B = \{ f(\mathbf{\tilde{I}}_i; \theta) \}_{i=1}^{B}, \quad \mathbf{\tilde{I}}_i \sim X_{\text{train}} \) \Comment{Forward pass}
            \vspace{3pt}

            \State \(\mathcal{L}_{L1} \gets \frac{1}{HWC} \sum_{i=1}^{HWC} \left|\mathbf{I_{\text{GT}}}_i - \mathbf{I}_{B_{i}}\right|\) \Comment{Compute the reconstruction loss}
            \vspace{2pt}

            \State \(\nabla_{\theta} \mathcal{L}_{L1} \gets \frac{\partial \mathcal{L}_{L1}}{\partial \theta}\) \Comment{Backward pass}
            \vspace{2pt}
        
            \State \(\nabla_{\theta} \mathcal{L}_{L1} \gets m \odot \nabla_{\theta} \mathcal{L}_{L1}\) \Comment{Mask gradients of pruned weights}
            
            \State \(\theta \gets \theta - \eta_t \,\nabla_\theta \mathcal{L}_{L1}\) \Comment{Parameter update via Adam}
            \vspace{2pt}
            
            \State \(\theta \gets m \odot \theta \) \Comment{Apply sparsity mask to updated weights}
        \EndFor
    \EndFor

    \State Determine pruning threshold \(\tau\) as the \(p\)-th percentile of \(\,|m \odot \theta|\), i.e., \( \tau = \operatorname{Quantile}_{p}\!\bigl(|m \odot \theta|\bigr) \)
    \State \( m' \gets \mathbb{1}(|m \odot \theta| \,\geq\, \tau)\) \Comment{Calculate new mask}
    \State \( \theta \gets m' \odot \theta_0 \) \Comment{Prune and reset to initial values}

    \State \( m \gets m' \) \Comment{Update mask}
\EndWhile
\State \Return Final sparse model \( f(\mathbf{\tilde{I}}; \theta) \)
\end{algorithmic}
\end{algorithm}

\subsection{MIR-L Optimization Algorithm}
The proposed MIR-L model optimizes the multi-task image restoration model (Section~\ref{sec:RestorationModel}) and prunes it with the LTH to obtain a sparse yet equally or more effective multi-task image restoration model. MIR-L, optimized with an \( L_1 \) reconstruction loss, is iteratively trained and pruned, until the target sparsity level is reached. Algorithm \ref{alg:promptir_lth} provides a formal outline of the model optimization and pruning process. Firstly, the dense network parameters and a binary mask are initialized (lines~1--2). Next, the network is trained for \(j\) epochs with a learning rate schedule that includes linear warmup followed by cosine annealing (lines~4--14). After each training step's backward pass, the gradients and weights are masked, to ensure they remain zeroed. After each training cycle, a pruning threshold \(\tau\) is determined based on a pruning rate \(p\), the mask is updated, the network is pruned and remaining weights are reset to their initial values (lines~15--18). This procedure is repeated until the target sparsity \(S\) is reached, resulting in a final sparse model.

Note that for layer-wise pruning, the threshold \(\tau\) is determined for each layer independently, whereas for global pruning \(\tau\) is determined across all layers.

\section{Experimental Evaluation}\label{sec:Experiments}
\subsection{Datasets}\label{subsec:Datasets}
We evaluate our MIR-L model, as well as the baselines following the evaluation protocol of~\cite{AdaIR,AirNet,PromptIR}. Specifcially, for image denoising, we use a combination of the BSD400~\cite{bsd400_ref} and WED~\cite{wed_ref} datasets for training. BSD400 consists of 400 training images and the WED dataset consists of 4,744 images. Due to training resourse constraints, we randomly selected 5\% of the images of each dataset for training. From the clean images, we generate the noisy images by adding Gaussian noise with different noise levels \(\sigma \in \{15, 25, 50\}\). Testing is performed on the Color BSD68~\cite{bsd68_ref} and Urban100~\cite{urban100_ref} datasets consisting of 68 and 100 images, respectively.
For image deraining, we use the Rain100L~\cite{rain100l_ref} dataset, which consists of 200 rainy-clean image pairs for training and 100 rainy-clean image pairs for testing. We randomly selected 10\% of the original pairs for training.
For image dehazing, we use the OTS~\cite{sots_ots_ref} dataset for training, which consists of 72,135 images. We randomly selected 3\% of the original pairs. Testing is performed on the SOTS~\cite{sots_ots_ref} dataset, consisting of 500 hazy-clean image pairs.
In the all-in-one setting (covering both training and testing), we combine the aforementioned datasets across denoising, deraining, and dehazing. This approach enables a unified evaluation of our method under a single model across multiple restoration tasks. All the datasets are publicly available for reproducibility purposes at https://github.com/Thomkat/MIR-L.

\subsection{Evaluation Protocol}
To evaluate the performance of our model, we need to specify appropriate metrics that objectively compare different models. In image restoration tasks, Peak Signal-to-Noise Ratio (PSNR) and Structural Similarity Index Measure (SSIM) are commonly used to assess the quality of restored images~\cite{MSPFN_Derain,AirNet,PromptIR,EPDN_Dehaze,Restormer,MPRNet,FFDNet_Denoise}. These metrics provide insight into the reconstruction fidelity and perceptual similarity of the restored images.

\subsubsection{Peak Signal-to-Noise Ratio (PSNR)} measures the ratio between the maximum possible power of a signal and the power of the noise that affects its representation. A higher PSNR value indicates better image quality, as it implies a lower level of distortion in the restored image. The PSNR is calculated as follows:

    \begin{equation}
        PSNR = 10 \cdot \log \left(\frac{MAX^2}{MSE} \right)
    \end{equation}

    \noindent where \textit{MAX} is the maximum possible pixel value, i.e., 255 for an 8-bit image and \textit{MSE} (Mean Squared Error) represents the average squared differences between corresponding pixels of the original and restored images.

\subsubsection{Structural Similarity Index Measure (SSIM)} quantifies the perceived visual quality of an image by considering structural information, luminance, and contrast. A higher SSIM value indicates better perceptual quality and structural similarity to the reference image. The SSIM is calculated as follows:

    \begin{equation}
        SSIM(x, y) = \frac{(2\mu_x \mu_y + C_1)(2\sigma_{xy} + C_2)}{(\mu_x^2 + \mu_y^2 + C_1)(\sigma_x^2 + \sigma_y^2 + C_2)}
    \end{equation}

\noindent where \( \mu_x \) and \( \mu_y \) are the mean intensities of images x and y, \( \sigma_x^2 \) and \( \sigma_y^2 \) are their variances, \( \sigma_{xy} \) is the covariance, and \( C_1 \) and \( C_2 \) are small constants to avoid instability.

While PSNR is useful for measuring absolute reconstruction fidelity, SSIM aligns better with human visual perception. Therefore, both PSNR and SSIM provide complementary insights into the performance of our model.

\subsection{Experimental Setup}
\subsubsection{Implementation Details}
All the experiments were performed on the NVIDIA A40 GPU, using PyTorch version 2.5.1. The model was trained for 120 epochs (15 warmup epochs) with a batch size of 8. Optimization was performed using the Adam optimizer with an \(L_1\) loss function and a learning rate of $2 \times 10^{-4}$. The target sparsity level \(S\) is 90\%, which corresponds to 15 pruning steps, and pruning rate \(p\) was set to 20\%. During training, the input images were randomly cropped into patches of size $64 \times 64$. To improve generalization, random horizontal and vertical flips were applied to the training data. Smaller datasets were artificially expanded by duplicating their images, while random augmentations ensured variation, allowing the model to perceive them as distinct and maintain a balanced training process.

\subsubsection{Examined Models}

\begin{table}[!t]
\caption{Examined Image Restoration Models}
\label{tab:exam_models}
\centering
\begingroup
\setlength{\tabcolsep}{3.9pt}
\resizebox{\columnwidth}{!}{
\begin{tabular}{l c c c c c c}
\toprule
Model & Single-Task & Multi-Task & \multicolumn{3}{c}{Task} & Sparse \\
\cmidrule(lr){4-6}
      &             &            & Deraining & Dehazing & Denoising & \\
\midrule
MSPFN~\cite{MSPFN_Derain}   & \checkmark &           & \checkmark &          &          &          \\
EPDN~\cite{EPDN_Dehaze}     & \checkmark &           &          & \checkmark &          &          \\
FFDNet~\cite{FFDNet_Denoise}& \checkmark &           &         &          & \checkmark &          \\
AirNet~\cite{AirNet}        &          & \checkmark & \checkmark & \checkmark & \checkmark &          \\
Restormer~\cite{Restormer}  &  &     \checkmark      &    \checkmark      &   \checkmark       &    \checkmark      &          \\
MPRNet~\cite{MPRNet}        &          & \checkmark & \checkmark & \checkmark & \checkmark &          \\
AdaIR~\cite{AdaIR}          &          & \checkmark & \checkmark & \checkmark & \checkmark &          \\
PromptIR~\cite{PromptIR}    &          & \checkmark & \checkmark & \checkmark & \checkmark &          \\
PIR-OSM I                  &  & \checkmark          & \checkmark & \checkmark & \checkmark & \checkmark \\
PIR-OSM II                 &  & \checkmark          & \checkmark & \checkmark & \checkmark & \checkmark \\
PIR-OSR I                  &  & \checkmark          & \checkmark & \checkmark & \checkmark & \checkmark \\
PIR-OSR II                 &  & \checkmark          & \checkmark & \checkmark & \checkmark & \checkmark \\
MIR-L-LW                  &  & \checkmark          & \checkmark & \checkmark & \checkmark & \checkmark \\
MIR-L-G                   &  & \checkmark          & \checkmark & \checkmark & \checkmark & \checkmark \\
\bottomrule
\end{tabular}
}
\endgroup
\end{table}

\begin{itemize}
    \item \textbf{MSPFN}\footnote{\url{https://github.com/kuijiang94/MSPFN}}~\cite{MSPFN_Derain}: A multi-scale progressive fusion network for image deraining, using cross-scale and intra-scale information with recurrent refinement.
    \item \textbf{EPDN}\footnote{\url{https://github.com/ErinChen1/EPDN}}~\cite{EPDN_Dehaze}: An enhanced Pix2pix Dehazing Network reframing dehazing as image-to-image translation, with a GAN-based enhancer module.
    \item \textbf{AirNet}\footnote{\url{https://github.com/XLearning-SCU/2022-CVPR-AirNet}}~\cite{AirNet}: An all-in-one image restoration network for unknown degradations via contrastive-based encoding and degradation-guided recovery.
    \item \textbf{Restormer}\footnote{\url{https://github.com/swz30/Restormer}}~\cite{Restormer}: A transformer-based restoration network for high-resolution images, utilizing attention for long-range dependencies.
    \item \textbf{FFDNet}\footnote{\url{https://github.com/cszn/FFDNet}}~\cite{FFDNet_Denoise}: A convolutional neural network (CNN) for image denoising using downsampled sub-images and a tunable noise-level map for spatially varying noise.
    \item \textbf{MPRNet}\footnote{\url{https://github.com/swz30/MPRNet}}~\cite{MPRNet}: A multi-stage all-in-one image restoration network that progressively refines spatial details.
    \item \textbf{AdaIR}\footnote{\url{https://github.com/c-yn/AdaIR}}~\cite{AdaIR}: An adaptive all-in-one image restoration network that mines low- and high-frequency features and modulates them bidirectionally for progressive correction.
    \item \textbf{PromptIR}\footnote{\url{https://github.com/va1shn9v/PromptIR}}~\cite{PromptIR}: An all-in-one blind image restoration model that generalizes to various unknown degradation types and levels by using prompt-based learning to encode degradation-specific information, dynamically guiding the restoration network.
    \item \textbf{PIR-OSM I}: A pruned version of the model described in Section~\ref{sec:RestorationModel} (one-shot, magnitude-based), obtained by removing 30\% of the smallest weights post-training and fine-tuning for an additional 5\% of training epochs.
    \item \textbf{PIR-OSM II}: A variant of PIR-OSM, removing 70\% of the smallest weights.
    \item \textbf{PIR-OSR I}: A pruned version of the model described in Section~\ref{sec:RestorationModel} (one-shot, random), obtained by randomly removing 30\% of weights post-training and fine-tuning for an additional 5\% of training epochs.
    \item \textbf{PIR-OSR II}: A variant of PIR-OSR, randomly removing 70\% of the weights.
    \item \textbf{MIR-L-LW (Layer-wise Pruning)}: Our proposed model based on the LTH with layer-wise pruning.
    \item \textbf{MIR-L-G (Global Pruning)}: Our proposed model based on the LTH with global pruning.
\end{itemize}

\noindent
In Table~\ref{tab:exam_models} we present an overview of the examined image restoration models. To ensure a fair comparison, we retrain all the aforementioned models using their publicly available implementations and the datasets described in Section~\ref{subsec:Datasets}. All the models are trained with an input patch size of \(64 \times 64\).

\subsection{Experimental Results}

\begin{table*}[!t]
\centering
\caption{Comparison of single-task results for (a) deraining, (b) dehazing, and (c) denoising. The best results are shown in bold, and the second-best are underlined. Our MIR-L-LW and MIR-L-G models drastically reduce trainable parameters while reaching performance similar to dense baseline models.}\label{tab:single}
\setlength{\tabcolsep}{10pt}
\begin{minipage}{0.49\textwidth}
\centering
\textbf{(a) Derain Model (Rain100L)}
\begin{tabular}{@{}lcc@{}}
\toprule
Method & PSNR/SSIM & \makecell{Trainable \\ Parameters} \\ 
\midrule
MSPFN~\cite{MSPFN_Derain} & 25.85/0.8118 & 21M \\
AirNet~\cite{AirNet} & 28.77/0.8867 & \underline{7.6M} \\
Restormer~\cite{Restormer} & 30.09/0.9114 & 26.1M \\ 
PromptIR~\cite{PromptIR} & \textbf{35.13}/\textbf{0.9683} & 35.6M \\
PIR-OSM I. & \underline{34.75}/0.9640 & 25.6M\\
PIR-OSM II. & 25.52/0.8140 & 12.4M\\
PIR-OSR I. & 25.74/0.8158 & 25.6M\\
PIR-OSR II. & 25.69/0.8142 & 12.4M\\
\midrule
MIR-L-LW & 32.14/0.9395 & \textbf{4.7M} \\
MIR-L-G & 34.72/\underline{0.9652} & \textbf{4.7M} \\
\bottomrule
\end{tabular}
\end{minipage}
\hfill
\begin{minipage}{0.49\textwidth}
\centering
\textbf{(b) Dehaze Model (SOTS)}
\begin{tabular}{@{}lcc@{}}
\toprule
Method & PSNR/SSIM & \makecell{Trainable \\ Parameters} \\ 
\midrule
EPDN~\cite{EPDN_Dehaze} & 24.57/0.9367 & 22.9M \\
AirNet~\cite{AirNet} & 22.13/0.9228 & \underline{7.6M} \\
Restormer~\cite{Restormer} & 25.32/0.9432 & 26.1M \\
PromptIR~\cite{PromptIR} & \underline{26.76}/\underline{0.9556} & 35.6M \\
PIR-OSM I. & 26.55/0.9525 & 25.6M\\
PIR-OSM II. & 18.75/0.8612 & 12.4M\\
PIR-OSR I. & 20.70/0.8838 & 25.6M\\
PIR-OSR II. & 17.25/0.8282 & 12.4M\\
\midrule
MIR-L-LW & 26.53/0.9533 & \textbf{4.7M} \\
MIR-L-G & \textbf{27.62}/\textbf{0.9609} & \textbf{4.7M} \\
\bottomrule
\end{tabular}
\end{minipage}

\vspace{8pt}

\begin{minipage}{\textwidth}
\centering
\textbf{(c) Denoise Model (BSD68 \& Urban100)}
\centering
\setlength{\tabcolsep}{12pt}
\begin{tabular}{@{}lcccccc@{}}
\toprule
Dataset & Method & \makecell{$\sigma=15$ \\ PSNR/SSIM} & \makecell{$\sigma=25$ \\ PSNR/SSIM} & \makecell{$\sigma=50$ \\ PSNR/SSIM} & \makecell{Trainable \\ Parameters} \\ 
\midrule
\multirow{10}{*}{BSD68} 
& FFDNet~\cite{FFDNet_Denoise} & 33.42/0.9240 & 30.93/0.8768 & 27.81/0.7838 & \textbf{494K} \\
& AirNet~\cite{AirNet} & \underline{33.89}/\underline{0.9324} & 31.28/\textbf{0.8883} & \underline{28.09}/\textbf{0.7997} & 7.6M \\
& Restormer~\cite{Restormer} & 33.64/0.9243 & 31.22/0.8796 & \textbf{28.12}/0.7896 & 26.1M \\ 
& PromptIR~\cite{PromptIR} & \textbf{33.97}/\textbf{0.9330} & \textbf{31.32}/\underline{0.8876} & 28.08/0.7961 & 35.6M \\
& PIR-OSM I. & 33.72/0.9272 & 31.07/0.8777 & 27.97/0.7873 & 25.6M\\
& PIR-OSM II. & 27.73/0.7058 & 23.54/0.5287 & 17.88/0.2980 & 12.4M\\
& PIR-OSR I. & 29.14/0.839 & 28.06/0.7848 & 24.61/0.5993 & 25.6M\\
& PIR-OSR II. & 25.44/0.6159 & 21.15/0.4397 & 15.59/0.2344 & 12.4M\\
\cmidrule(lr){2-6}
& MIR-L-LW & 33.01/0.9208 & 30.40/0.8688 & 27.06/0.7554 & \underline{4.7M} \\
& MIR-L-G & 33.85/0.9298 & \underline{31.30}/0.8862 & 28.07/\underline{0.7962} & \underline{4.7M} \\
\midrule
\midrule
\multirow{10}{*}{Urban100} 
& FFDNet~\cite{FFDNet_Denoise} & 32.65/0.9316 & 30.57/0.9017 & 27.51/0.8367 & \textbf{494K} \\
& AirNet~\cite{AirNet} & \underline{34.30}/\textbf{0.9476} & \underline{31.99}/\textbf{0.9219} & \underline{28.72}/\textbf{0.8661} & 7.6M \\
& Restormer~\cite{Restormer} & \textbf{34.36}/\underline{0.9449} & \textbf{32.05}/\underline{0.9183} & \textbf{28.83}/\underline{0.8608} & 26.1M \\ 
& PromptIR~\cite{PromptIR} & 33.90/0.9433 & 31.51/0.9139 & 28.23/0.8522 & 35.6M \\
& PIR-OSM I. & 33.46/0.9363 & 31.10/0.9049 & 28.04/0.8442 & 25.6M\\
& PIR-OSM II. & 27.61/0.7262 & 23.54/0.5750 & 17.95/0.3696 & 12.4M\\
& PIR-OSR I. & 27.08/0.8389 & 26.31/0.7858 & 23.57/0.616 & 25.6M\\
& PIR-OSR II. & 25.52/0.6500 & 21.28/0.4998 & 15.75/0.3080 & 12.4M\\
\cmidrule(lr){2-6}
& MIR-L-LW & 32.00/0.9250 & 29.54/0.8846 & 26.04/0.7910 & \underline{4.7M} \\
& MIR-L-G & 33.71/0.9392 & 31.50/0.9131 & 28.26/0.8529 & \underline{4.7M} \\
\bottomrule
\end{tabular}
\end{minipage}
\end{table*}

Table~\ref{tab:single} compares our MIR-L against conventional one-shot pruning baselines and model baselines on the following single-task settings: Table~\ref{tab:single}a reports deraining results on the Rain100L dataset, Table~\ref{tab:single}b reports dehazing results on the SOTS dataset and Table~\ref{tab:single}c reports denoising results on the BSD68 and Urban100 datasets. In the single task setting, separate models are trained for each individual degradation (Table \ref{tab:exam_models}). Multi-task models have a higher number of trainable parameters than single-task models but they achieve better restoration performance. Although one-shot magnitude (PIR-OSM~I.~\&~II.) and random pruning (PIR-OSR~I.~\&~II.) reduce the trainable parameters, they show a steep drop in performance at high sparsity levels, expressed by fewer trainable parameters. The proposed MIR-L-LW and MIR-L-G drastically reduce the parameters, down to 4.7M, while preserving the performance high. Our strategy achieves superior performance by gradually pruning the model and resetting the remaining weights to their original values. This process allows the optimization to relearn the weights and recover any lost performance by modifying the relationships between the surviving weights. In subsequent rounds, less important weights are pruned again, ensuring that the most critical parts of the network are preserved. By contrast, conventional one-shot pruning methods remove a large portion of weights all at once, leaving little opportunity for the model to adjust and fully recover the lost performance. MIR-L-G consistently outperforms MIR-L-LW in all settings, demonstrating that global pruning more effectively discovers winning tickets in large networks.

Table~\ref{tab:aio} compares our MIR-L against conventional one-shot pruning baselines and model baselines on the multi-task setting: Table~\ref{tab:aio}a reports deraining results on the Rain100L dataset, Table~\ref{tab:aio}b reports dehazing results on the SOTS dataset and Table~\ref{tab:aio}c reports denoising results on the BSD68 and Urban100 datasets. In the multi-task (all-in-one) setting, a model is trained to simultaneously handle multiple degradations. Similarly to the single-task settings, one-shot magnitude (PIR-OSM~I.~\&~II.) and random pruning (PIR-OSR~I.~\&~II.) reduce the parameters, while their performance degrades significantly as sparsity increases. The proposed MIR-L-LW and MIR-L-G achieve greater performance than PIR-OSM and PIR-OSR, using only 4.7M parameters, an approximate 87\% reduction compared to the dense model's 35.6M parameters on average, corresponding to a compression rate of x7.57. Similarly to the single-task setting, MIR-L-G outperforms MIR-L-LW, with the former achieving restoration performance that reaches or exceeds state-of-the-art both in terms of PSNR and SSIM.

\begin{figure}[!t]
\includegraphics[width=0.5\textwidth]{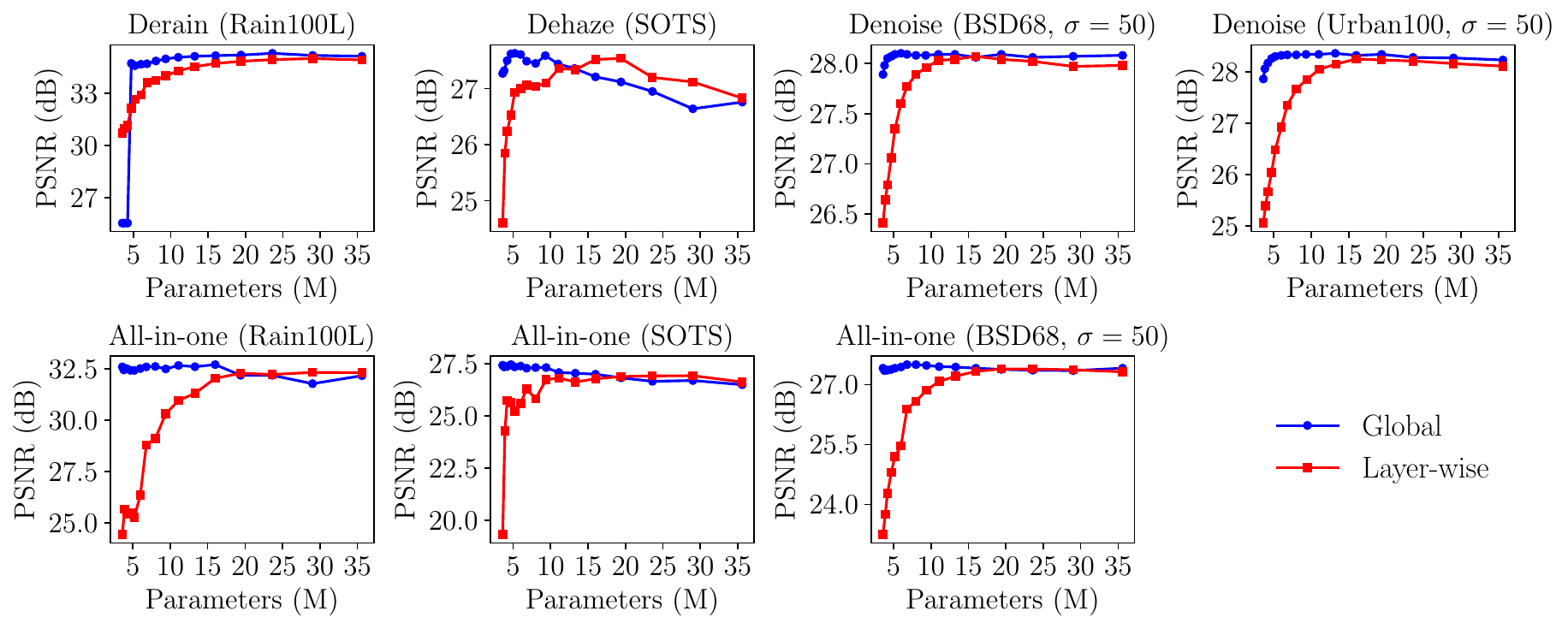}
\caption{PSNR vs. trainable parameter count across progressive pruning steps. The x-axis denotes the number of trainable parameters, where a larger pruning step corresponds to fewer trainable parameters. MIR-L-G consistently maintains high performance as step (sparsity) increases, while MIR-L-LW experiences a sharp drop at higher pruning levels.} \label{fig:steps}
\end{figure}

Figure~\ref{fig:steps} reports PSNR when varying the number of trainable parameters for layer-wise and global pruning. The x-axis indicates the number of trainable parameters, where a larger pruning step corresponds to fewer trainable parameters. We observe that MIR-L-G consistently outperforms MIR-L-LW as sparsity increases, primarily because global pruning selectively removes redundant weights across all layers, avoiding bottlenecks in thinner layers and thus preserving the subnetwork’s overall representational capacity. In the single-task settings, both pruning strategies initially maintain high PSNR values, but as pruning becomes more aggressive, layer-wise pruning shows a significant performance drop compared to global pruning. An exception is deraining, where global pruning shows a large drop at higher sparsity levels compared to layer-wise pruning. This occurs because weights essential for deraining performance are pruned by the global magnitude pruning criterion during these steps. In the multi-task (all-in-one) setting, we observe a similar trend: global pruning not only maintains a higher PSNR across all tasks, but performance improves in all tasks as parameters are reduced, whereas layer-wise pruning shows a steep performance drop at higher sparsity levels.

\begin{table*}[!t]
\centering
\caption{Comparison of multi-task (all-in-one) results for (a) deraining, (b) dehazing, and (c) denoising. The best results are shown in bold, and the second-best are underlined. Our MIR-LW and MIR-L-G models achieve performance similar to or higher than state-of-the-art, with substantially fewer trainable parameters than dense models.}\label{tab:aio}
\setlength{\tabcolsep}{18pt}
\begin{minipage}{0.49\textwidth}
\centering
\textbf{(a) Rain100L Dataset}
\begin{tabular}{@{}lcc@{}}
\toprule
Method & PSNR/SSIM & \makecell{Trainable \\ Parameters} \\ 
\midrule
MPRNet~\cite{MPRNet} & 27.64/0.8477 & 39.5M \\
AirNet~\cite{AirNet} & 27.83/0.8809 & \underline{7.6M} \\ 
PromptIR~\cite{PromptIR} & \underline{32.17}/\underline{0.9372} & 35.6M \\
AdaIR~\cite{AdaIR} & 25.90/0.8409 & 28.8M \\
PIR-OSM I. & 31.85/0.9272 & 25.6M\\
PIR-OSM II. & 26.59/0.8365 & 12.4M\\
PIR-OSR I. & 25.39/0.8101 & 25.6M\\
PIR-OSR II. & 26.02/0.8171 & 12.4M\\
\midrule
MIR-L-LW & 25.49/0.8125 & \textbf{4.7M} \\
MIR-L-G & \textbf{32.43}/\textbf{0.9425} & \textbf{4.7M} \\
\bottomrule
\end{tabular}
\end{minipage}
\hfill
\begin{minipage}{0.49\textwidth}
\centering
\textbf{(b) SOTS Dataset}
\begin{tabular}{@{}lcc@{}}
\toprule
Method & PSNR/SSIM & \makecell{Trainable \\ Parameters} \\ 
\midrule
MPRNet~\cite{MPRNet} & 24.34/0.9350 & 39.5M \\
AirNet~\cite{AirNet} & 22.41/0.8738 & \underline{7.6M} \\ 
PromptIR~\cite{PromptIR} & 26.49/0.9535 & 35.6M \\
AdaIR~\cite{AdaIR} & \underline{27.09}/\underline{0.9575} & 28.8M \\
PIR-OSM I. & 26.43/0.9524 & 25.6M\\
PIR-OSM II. & 17.54/0.8399 & 12.4M\\
PIR-OSR I. & 18.02/0.8404 & 25.6M\\
PIR-OSR II. & 16.54/0.8180 & 12.4M\\
\midrule
MIR-L-LW & 25.63/0.9446 & \textbf{4.7M} \\
MIR-L-G & \textbf{27.45}/\textbf{0.9591} & \textbf{4.7M} \\
\bottomrule
\end{tabular}
\end{minipage}

\vspace{8pt}

\begin{minipage}{\textwidth}
\centering
\textbf{(c) BSD68 Dataset}\par
\setlength{\tabcolsep}{10pt}
\begin{tabular}{@{}lcccc@{}}
\toprule
Method & \makecell{$\sigma=15$ \\ PSNR/SSIM} & \makecell{$\sigma=25$ \\ PSNR/SSIM} & \makecell{$\sigma=50$ \\ PSNR/SSIM} & \makecell{Trainable \\ Parameters} \\ 
\midrule
MPRNet~\cite{MPRNet} & 32.15/0.8976 & 30.10/0.8552 & 27.36/0.7647 & 39.5M \\
AirNet~\cite{AirNet} & 32.79/0.9167 & 30.30/0.8602 & 27.07/0.7437 & \underline{7.6M} \\ 
PromptIR~\cite{PromptIR} & 33.50/0.9247 & 30.79/0.8734 & 27.41/0.7667 & 35.6M \\
AdaIR~\cite{AdaIR} & \underline{33.52}/\underline{0.9250} & \underline{30.82}/\underline{0.8747} & \textbf{27.48}/\underline{0.7695} & 28.8M \\
PIR-OSM I. & 32.97/0.9148 & 30.29/0.8556 & 26.76/0.7202 & 25.6M\\
PIR-OSM II. & 25.91/0.6374	& 21.66/0.4605 & 16.06/0.2475 & 12.4M\\
PIR-OSR I. & 25.38/0.6382	& 21.47/0.4628 & 16.00/0.2504 & 25.6M\\
PIR-OSR II. & 24.62/0.5965	& 20.56/0.4218 & 15.14/0.2220 & 12.4M\\
\midrule
MIR-L-LW & 31.22/0.8731 & 28.55/0.7946 & 24.80/0.6166 & \textbf{4.7M} \\
MIR-L-G & \textbf{33.53}/\textbf{0.9269} & \textbf{30.83}/\textbf{0.8772} & \textbf{27.48}/\textbf{0.7736} & \textbf{4.7M} \\
\bottomrule
\end{tabular}
\end{minipage}
\end{table*}

\section{Conclusion}\label{sec:Conclusion} This paper proposes a pruning strategy for multi-task image restoration models based on lottery tickets (MIR-L), focusing on the deraining, dehazing, and denoising tasks. To deal with the overparameterization of multi-task image restoration models, we presented an iterative pruning strategy that removes low-magnitude weights in multiple rounds, while resetting the surviving weights to their initial values. The proposed MIR-L optimization algorithm discovers sparse "winning tickets" capable of matching or surpassing the performance of their dense counterparts, at a fraction of trainable parameters. Our experiments demonstrated that MIR-L effectively reduces the number of trainable parameters by up to 90\% across both single-task and multi-task settings, while maintaining high performance on benchmark datasets. This model size reduction and low computational requirements are beneficial for web platforms, allowing faster delivery of high-quality images and improved user experience even on less powerful client devices.
In future work, exploring more sophisticated pruning criteria, such as SynFlow~\cite{synflow_related}, or expanding the implementation to image restoration tasks commonly used in real-time applications, such as super-resolution~\cite{ESRT_SISR,CVANet_SISR}, may offer further improvements in both efficiency and image restoration accuracy.

\bibliographystyle{IEEEtran}
\bibliography{references}

\end{document}